\documentclass{article}

\usepackage[final]{neurips_2020}

\usepackage[utf8]{inputenc} 
\usepackage[T1]{fontenc}    
\usepackage{hyperref}       
\usepackage{url}            
\usepackage{booktabs}       
\usepackage{amsfonts}       
\usepackage{nicefrac}       
\usepackage{microtype}      

\usepackage{graphicx}
\usepackage{multirow}

\usepackage{color}

\usepackage{amsmath}
\usepackage{amssymb}

\usepackage{amsthm}
\newtheorem{definition}{Definition}
\newtheorem{theorem}{Theorem}

\usepackage{algorithm}
\usepackage{algorithmic}

\newcommand{\RR}{\mathbb{R}}

\newcommand{\LL}{\mathcal{L}}
\newcommand{\stiefel}[2]{\operatorname{St}\left(#1, #2\right)}
\DeclareMathOperator{\st}{s.t.}
\newcommand{\norm}[1]{\left\lVert#1\right\rVert}

\newcommand{\rank}[1]{\operatorname{rank}\left(#1\right)}
\newcommand{\dimof}[1]{\operatorname{dim}\left(#1\right)}

\newcommand{\enc}[1]{enc\left(#1; \theta_e\right)}
\newcommand{\dec}[1]{dec\left(#1; \theta_d\right)}
\newcommand{\h}[1]{h\left(#1; \theta_h\right)}

\title{Learning Self-Expression Metrics for Scalable and Inductive Subspace Clustering}

\author{%
Julian Busch\footnote{Contact Author}\And
Evgeniy Faerman\And
Matthias Schubert\And
Thomas Seidl\And
\vspace{-1em}
\\
Ludwig-Maximilians-Universit\"at M\"unchen\\
Munich Center for Machine Learning (MCML)\\
\texttt{\{busch, faerman, schubert, seidl\}@dbs.ifi.lmu.de}
}

\begin{document}

\maketitle

Subspace clustering has established itself as a state-of-the-art approach to clustering high-dimensional data. In particular, methods relying on the self-expressiveness property have recently proved especially successful. 
However, they suffer from two major shortcomings: First, a quadratic-size coefficient matrix is learned directly, preventing these methods from scaling beyond small datasets. Secondly, the trained models are transductive and thus cannot be used to cluster out-of-sample data unseen during training.
Instead of learning self-expression coefficients directly, we propose a novel metric learning approach to learn instead a subspace affinity function using a siamese neural network architecture. 
Consequently, our model benefits from a constant number of parameters and a constant-size memory footprint, allowing it to scale to considerably larger datasets. In addition, we can formally show that out model is still able to exactly recover subspace clusters given an independence assumption.
The siamese architecture in combination with a novel geometric classifier further makes our model inductive, allowing it to cluster out-of-sample data. Additionally, non-linear clusters can be detected by simply adding an auto-encoder module to the architecture. The whole model can then be trained end-to-end in a self-supervised manner. This work in progress reports promising preliminary results on the MNIST dataset. In the spirit of reproducible research, me make all code publicly available. 
\setcounter{footnote}{0}\footnote{\url{https://github.com/buschju/sscn}}
In future work we plan to investigate several extensions of our model and to expand experimental evaluation.

\section{Introduction}
\label{sec:introduction}
\emph{Subspace clustering} \citep{vidal2011subspace} assumes the data to be sampled from a union of low-dimensional subspaces of the full data space. The goal is to recover these subspaces and to correctly assign each data point to its respective subspace cluster.
As a state-of-the-art approach to clustering high-dimensional data, it enables a multitude of applications, including image segmentation \citep{ma2007segmentation,yang2008unsupervised}, motion segmentation \citep{kanatani2001motion,elhamifar2009sparse,ji2016robust}, image clustering \citep{ho2003clustering,elhamifar2013sparse} and clustering gene expression profiles \citep{mcwilliams2014subspace}. For instance, face images of a subject under fixed pose and varying lighting conditions \citep{basri2003lambertian} or images of hand-written digits with different rotations, translations and other natural transformations \citep{hastie1998metrics} have been shown to lie in low-dimensional subspaces.

Recently, \emph{self-expressiveness}-based methods \citep{elhamifar2009sparse,liu2010robust,lu2012robust,elhamifar2013sparse,liu2013robust,wang2013provable,feng2014robust,ji2014efficient,vidal2014low,ji2015shape,you2016oracle} have proved especially successful. The main idea is that each point can be expressed by a linear combination of points from the same subspace. This property is used to learn a quadratic-size coefficient matrix from which cluster labels can be extracted in a post-processing step using spectral clustering. The quadratic number of parameters prevents these methods from scaling beyond small datasets and makes them transductive and thus inapplicable to out-of-sample data unseen during training. In contrast, our model requires only a constant number of parameters to provably provide the same expressive power. A classifier leveraging the unique geometric properties of our model further makes it inductive and enables it to cluster out-of-sample data.

In practice, we can usually not assume that the data exactly fits in linear subspaces, e.g., due to noise or additional non-linearities in the underlying data generating process which was not accounted for.
Instead, we might rather assume the data to be situated on different \emph{non-linear} sub-manifolds. Some methods \citep{chen2009spectral,patel2013latent,patel2014kernel,xiao2016robust,yin2016kernel,ji2017adaptive} rely on the \emph{kernel trick} to account for non-linearity. However, it is usually not clear whether a particular pre-defined kernel function is particularly suitable for subspace clustering. 
More recent methods \citep{peng2016deep,ji2017deep,zhou2018deep,zhang2019self,seo2019deep,kheirandishfard2020multi} learn a suitable feature transformation explicitly in an \emph{end-to-end} differentiable model. In particular, \emph{Deep Subspace Clustering Networks (DSC-Net)} \citep{ji2017deep} introduced the idea of modeling the coefficient matrix as a dense neural network layer, called self-expressive layer, and training it jointly with an auto-encoder. As a result, the encoder represents a feature transformation which has been optimized w.r.t. linear cluster structure in latent space. \emph{Self-Supervised Convolutional Subspace Clustering Networks (S$^2$ConvSCN)} \citep{zhang2019self} learn an additional classifier which can be applied to out-of-sample data but still rely on the full coefficient matrix and spectral clustering. Our model on the other hand offers both, applicability to out-of-sample data and scalability.

Several works have addressed the challenge of \emph{scalability} but are either only able to detect linear clusters \citep{you2016scalable,rahmani2017innovation}, rely on a $k$-means-like procedure which requires good initialization and is sensitive to outliers \citep{zhang2018scalable} or still fully parametrize coefficient matrices which need to be re-learned from scratch for each new data batch and come with no theoretical guarantees \citep{zhang2019neural}. In contrast, our model is suitable to detect non-linear clusters, can be trained end-to-end with back-propagation and provides a theoretical foundation.

In summary, we propose, to the best of our knowledge, the first metric learning approach to subspace clustering, which enables a quadratic reduction of the number of parameters and memory footprint compared to existing methods while maintaining theoretical performance guarantees. Our model is applicable to out-of-sample data, suitable to detect non-linear clusters and can be trained end-to-end with back-propagation. 

\begin{figure}[t]
\begin{center}
\includegraphics[width=\textwidth]{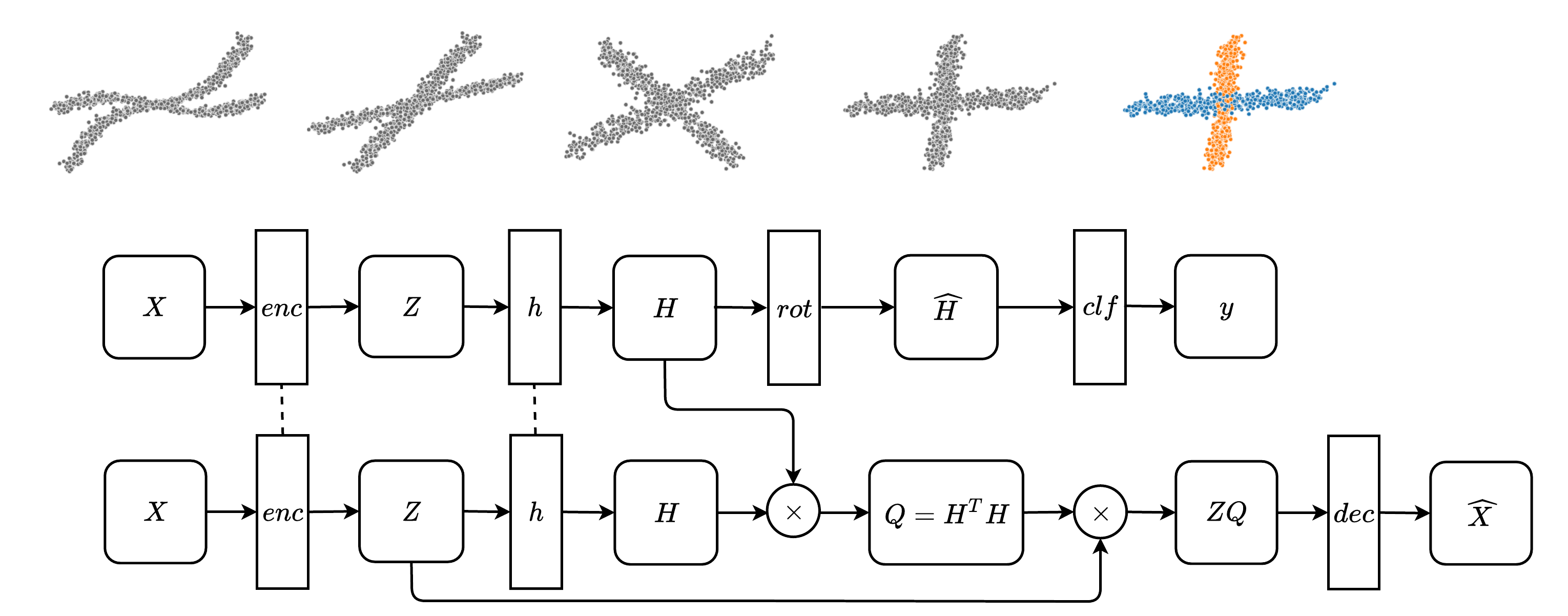}
\caption{Data flow within our model. Square boxes denote tensors, rectangular boxes denote functions, and dashed lines indicate parameter sharing.
After being mapped into a space in which the data fits better into independent linear subspaces using an encoder function $enc$, the data is mapped into a space using a function $h$ in which independent clusters are orthogonal and dot-products corresponds to self-expression coefficients. Afterwards, the data is rotated into pre-defined axis-aligned subspaces and assigned to clusters based on the orthogonal projection distance (upper path). The original data is reconstructed from the self-expressed data in latent space by a decoder function $dec$ to ensure that the learned embeddings are compatible with the original data (lower path).
}
\label{fig:model}
\end{center}
\end{figure}

\section{Siamese Subspace Clustering Networks}
\label{sec:method}
In subspace clustering, we are given a set set of points $\{x_i\}_{i=1}^N \subseteq \RR^{d_X}$ sampled from a union of subspaces $\{S_i\}_{i=1}^K$ of unknown dimensions and arranged as columns of a data matrix $X \in \RR^{d_X \times N}$. The goal is to recover these subspaces and to correctly assign each data point to its respective subspace cluster.
While there exist many different variants of self-expressive subspace clustering, we focus here on a relaxed noise-aware version of \emph{Efficient Dense Subspace Clustering (EDSC)} \citep{ji2014efficient}:

\begin{definition}[Efficient Dense Subspace Clustering \citep{ji2014efficient}]
\begin{equation}
    \min_{C \in \RR^{N \times N}} \frac{1}{2} \norm{C}_F^2 + \frac{\lambda}{2} \norm{X - XC}_F^2.
    \label{eq:noisy}
\end{equation}
\end{definition}

where the $i$-th column of the $N \times N$ coefficient matrix $C$ contains the coefficients for expressing $x_i$. Regularization of $C$ ensures that $x_i$ is expressed using only points from the same subspace. 
Given the learned coefficient matrix, cluster assignments can be extracted in a post-processing step by applying spectral clustering to the subspace affinity matrix $A = \left|C\right| + \left|C\right|^T$.
The unique solution to this problem can be expressed in closed-form as the solution $C^\ast$ of the linear system $\left( I + \lambda X^TX \right) C = \lambda X^TX$ \citep{ji2014efficient}. Let $r := \rank{X} = \dimof{\bigoplus_{i=1}^K S_i}$ denote the rank of $X$. In a noise-free setting, if the subspaces are \emph{independent}, i.e., if $ r = \sum_{i=1}^K \dimof{S_i}$, then $C^\ast$ is guaranteed to be block-diagonal with $C^\ast_{ij} = 0$ if $x_i$ and $x_j$ originate from different subspaces \citep{vidal2008multiframe}. The corresponding solution is called \emph{subspace-preserving}.

The central idea of our approach is to view subspace clustering from a metric learning perspective. To this end, we employ a siamese neural network \citep{bromley1994signature} consisting of two identical branches with shared weights and mirrored parameter updates which is optimized such that dot-products in latent space correspond to self-expression coefficients:

\begin{definition}[Siamese Dense Subspace Clustering]
\begin{align}
    \begin{split}
        \min_{\theta_h} & \quad \frac{1}{2} \norm{Q}_F^2 + \frac{\lambda}{2} \norm{X - XQ}_F^2 \\
        \st & \quad Q = H^T H, \quad H = \h{X}
    \end{split}
    \label{eq:learn_noisy}
\end{align}
\end{definition}

where $Q \in \RR^{N \times N}$ contains the self-expression coefficients corresponding to dot-products of the embeddings $H \in \RR^{d_H \times N}$ computed by the embedding function $h$. Note that weight sharing leads to symmetric coefficient matrices.
Even though the reduction of parameters compared to (\ref{eq:noisy}) is quadratic, we can show that this model is able to recover the exact solution to the original subspace clustering problem, even when $h$ consists of only a single linear layer with a sufficient number of neurons. Note that (\ref{eq:learn_noisy}) is convex in this case.

\begin{theorem}
\label{thm:frobenius}
Let $h(X) = WX$, $W \in \RR^{d_H \times d_X}$, $d_H \geq r$, then (\ref{eq:learn_noisy}) attains its global minimum at $W^\ast = R\sqrt{\lambda \left( I - \lambda \left( \Sigma_r^{-2} + \lambda I \right)^{-1} \right)} U_r^T$ where $X = U_r \Sigma_r V_r^T$ is the reduced SVD of $X$ and $R \in \stiefel{d_H}{r}$ is an arbitrary orthonormal matrix. The unique optimal coefficient matrix $Q^\ast$ of (\ref{eq:learn_noisy}) corresponds to the unique solution of (\ref{eq:noisy}).
\end{theorem}




Above, $\stiefel{n}{p} = \left\{ X \in \RR^{n \times p} \mid X^TX = I \right\}$ for $n \geq p$ denotes the \emph{Stiefel manifold} which is composed of all $n \times p$ orthonormal matrices. Since (\ref{eq:learn_noisy}) leads to a well-studied optimal solution, it can be analyzed directly within existing theory. In particular, it is guaranteed that under the independence assumption and in a noise-free setting, (\ref{eq:learn_noisy}) yields a subspace-preserving solution.
Also note that we don't need to know the exact rank of $X$, it is sufficient to have an upper bound. Since $r \leq \sum_{i=1}^K \text{dim}(S_i)$, we can simply estimate the number of clusters $K$ and the maximum cluster dimension $q$ and set $d_H = Kq$ and $R \in \stiefel{d_H}{d_H}$.

Since we can choose $R$ arbitrarily from $\stiefel{d_H}{d_H}$ and still obtain the same optimal coefficient matrix $Q^\ast$, we are able to optimize $R$ on the Stiefel manifold w.r.t. to a cluster assignment objective where we take advantage of the observation that points from independent clusters will have orthogonal embeddings in $H$. To this end, we compute rotated embeddings $\widehat{H} = RH$ and then classify points by assigning them to their closest subspace w.r.t. orthogonal projection distance and applying the \emph{softmin} function: $y_{ij} = \nicefrac{\exp{\left(-||\hat{h}_i - S_jS_j^T\hat{h}_i||_2^2\right)}}{\sum_{k=1}^K \exp{\left(-||\hat{h}_i - S_kS_k^T\hat{h}_i||_2^2\right)}}$. The subspaces are fixed a-priori to be axis-aligned and don't need to be optimized. The matrix $R$ is optimized such that classifications agree with self-expression affinities. For now, we compute the coefficient matrix of the training set using our trained model and then apply the same post-processing as in \citep{ji2017deep} to obtain pseudo-labels which are used to train the classifier using \emph{cross-entropy} loss and the \emph{Cayley-Adam} algorithm \citep{li2020efficient}. In future work, we plan to employ a triplet-loss \citep{hermans2017defense} which does not rely on spectral clustering and to additionally optimize the remaining model parameters with feedback from the classifier.

To account for non-linearity, we can simply add an auto-encoder to our model with the task of mapping the original data into a $d_Z$-dim. latent space in which the data better fits in linear subspaces and additionally the independence assumption can be better satisfied. This non-linear transformation is learned together with the rest of the model. We formalize our complete model in Definition \ref{def:sscn}.

\begin{definition}[Siamese Subspace Clustering Network (SSCN)]
\begin{align}
    \begin{split}
        \min_{\theta_e, \theta_d, \theta_h, R} & \quad \frac{1}{2} \norm{Q}_F^2 + \frac{\lambda_1}{2} \norm{Z - Z Q}_F^2 + \frac{\lambda_2}{2} \norm{X - \widehat{X}}_F^2 + \lambda_3 \LL_{clf}(R; X) \\
        \st & \quad Q = H^T H, \quad H = \h{Z},\\
        & \quad Z = \enc{X}, \quad \widehat{X} = \dec{Z Q} \\
    \end{split}
    \label{eq:sscn}
\end{align}
\label{def:sscn}
\end{definition}

Above, $Z \in \RR^{d_Z \times N}$ are non-linear embeddings of the input $X$ computed by the encoder function $enc$.
After self-expression in latent space, $X$ is reconstructed as $\widehat{X} \in \RR^{d_X \times N}$ using $dec$, a decoder function matching $enc$. The reconstruction loss ensures that the learned embeddings are actually compatible with the original data and prevents trivial solutions. 
Note that training with the full data batch $X$ would lead to materialization of the full $N \times N$ coefficient matrix $Q$. This is not an issue for our model, however, since it can be trained with mini-batches and thus scale to large datasets. The only requirements are that batches need to be sampled uniformly at random and that the batch-size needs to be sufficiently large so that we sample enough instances from each class on average and thus obtain a representative sample.
An illustration of the data flow is provided in Figure \ref{fig:model}.

\section{Experiments}
\label{sec:experiments}
\begin{table*}[t]
	\centering
	\resizebox{\textwidth}{!}{%
\begin{tabular}{lrrrrr}
    \toprule
     & \textbf{ACC} & \textbf{ARI} & \textbf{NMI} & \textbf{\#Parameters} & \textbf{GPU-Memory (GB)} \\
     \midrule
     \textbf{DSC-Net} & $63.54 \pm 0.00$ & $57.42 \pm 0.00$ & $\mathbf{72.34 \pm 0.00}$ & $100,014,991 \phantom{\;\mathbf{(-99.93\%)}}$ & $2.71 \phantom{\;\mathbf{(-92.96\%)}}$ \\
     \textbf{SSCN} & $\mathbf{67.98 \pm 3.40}$ & $\mathbf{58.53 \pm 3.34}$ & $69.48 \pm 2.38$ & $\mathbf{66,291 \;(-99.93\%)}$ & $\mathbf{0.19 \;(-92.96\%)}$ \\
     \midrule
     \textbf{SSCN-OoS} & $67.39 \pm 3.38$ & $57.10 \pm 3.27$ & $67.16 \pm 2.34$ & $66,291 \phantom{\;\mathbf{(-99.93\%)}}$ & $0.19 \phantom{\;\mathbf{(-92.96\%)}}$ \\
     \bottomrule
\end{tabular}
	}
	\caption{Results on the MNIST dataset. Upper part: Transductive clustering of the 10,000 test images. Lower part: Inductive clustering of the 60,000 out-of-sample training images using our previously trained model. Note that our model did not see these images during training and that DSC-Net does not support clustering out-of-sample data and would require more than 4.9B parameters and 39GB of GPU-memory to cluster the whole dataset. All results are aggregated over 10 independent runs with different random initializations. For better comparability, all models use the same pre-trained auto-encoder. Results for DSC-Net exhibit no variation since the model uses constant initialization.}
	\label{tab:results}
\end{table*}

As a first proof of concept, we compare our model with \emph{DSC-Net} \citep{ji2017deep} on the \emph{MNIST} dataset \citep{lecun1998mnist}. By default, we use a small convolutional auto-encoder and the same parameter settings for both models wherever possible. For SSCN, we model $h$ as a single linear layer without bias as motivated above and train with a batch-size of 1000. The results are summarized in Figure \ref{tab:results}. We can see that our model provides competitive performance while drastically reducing the required number of model parameters and GPU-memory. Even large amounts of out-of-sample data can be clustered reliably without any memory overhead.
All hyper-parameter values and complete code for reproducing the reported results are provided in our public code repository. 
In future work we plan to train our model end-to-end using a triplet-loss and additional feedback from the classifier to the encoder and self-expression module. We further plan to evaluate on more datasets, with different architectural choices and against more baselines.

\begin{ack}
This work has been partially funded by the German Federal Ministry of Education and Research (BMBF) under Grant No. 01IS18036A. The authors of this work take full responsibilities for its content.
\end{ack}

\bibliographystyle{plainnat}
\bibliography{bibliography}

\end{document}